\documentclass[conference]{IEEEtran}
\usepackage{times}

\usepackage[numbers]{natbib}
\usepackage{multicol}
\usepackage[bookmarks=true]{hyperref}
\usepackage{fancyvrb}
\usepackage{graphicx}
\usepackage{float}

\newtheorem{definition}{Definition}

\begin{document}

\newcommand{\JB}[1]{\textcolor{orange}{(Jaebok) #1}}

% paper title
\title{Planning Based System for Child-Robot Interaction in Dynamic Play Environments}

% You will get a Paper-ID when submitting a pdf file to the conference system
\author{Author Names Omitted for Anonymous Review. Paper-ID [add your ID here]}

\author{\authorblockN{Vicky Charisi}
\authorblockA{Human Media Interaction\\
University of Twente\\
%Atlanta, Georgia 30332--0250\\
Email: 	v.charisi@utwente.nl}
\and
\authorblockN{Bram Ridder}
\authorblockA{Department of Informatics\\ 
King's College London\\
Email: bernardus.ridder@kcl.ac.uk}
\and
\authorblockN{Jaebok Kim}
\authorblockA{Human Media Interaction\\
University of Twente\\
%Atlanta, Georgia 30332--0250\\
Email: 	j.kim@utwente.nl}
\and
\authorblockN{Vanessa Evers}
\authorblockA{Human Media Interaction\\
University of Twente\\
%Atlanta, Georgia 30332--0250\\
Email: 	v.evers@utwente.nl}
}

% avoiding spaces at the end of the author lines is not a problem with
% conference papers because we don't use \thanks or \IEEEmembership

% for over three affiliations, or if they all won't fit within the width
% of the page, use this alternative format:
% 
%\author{\authorblockN{Michael Shell\authorrefmark{1},
%Homer Simpson\authorrefmark{2},
%James Kirk\authorrefmark{3}, 
%Montgomery Scott\authorrefmark{3} and
%Eldon Tyrell\authorrefmark{4}}
%\authorblockA{\authorrefmark{1}School of Electrical and Computer Engineering\\
%Georgia Institute of Technology,
%Atlanta, Georgia 30332--0250\\ Email: mshell@ece.gatech.edu}
%\authorblockA{\authorrefmark{2}Twentieth Century Fox, Springfield, USA\\
%Email: homer@thesimpsons.com}
%\authorblockA{\authorrefmark{3}Starfleet Academy, San Francisco, California 96678-2391\\
%Telephone: (800) 555--1212, Fax: (888) 555--1212}
%\authorblockA{\authorrefmark{4}Tyrell Inc., 123 Replicant Street, Los Angeles, California 90210--4321}}

\maketitle

\begin{abstract}
This paper describes the initial steps towards the design of a robotic system that intends to perform actions autonomously in a naturalistic play environment. At the same time it aims for social human-robot interaction~(HRI), focusing on children. We draw on existing theories of child development and on dimensional models of emotions to explore the design of a dynamic interaction framework for natural child-robot interaction. In this dynamic setting, the social HRI is defined by the ability of the system to take into consideration the socio-emotional state of the user and to plan appropriately by selecting appropriate strategies for execution. The robot needs a temporal planning system, which combines features of task-oriented actions and principles of social human robot interaction. We present initial results of an empirical study for the evaluation of the proposed framework in the context of a collaborative sorting game.
\end{abstract}

\IEEEpeerreviewmaketitle

\section{Introduction}

The capabilities of autonomous robotic systems have increased significantly over the last few years and are used in increasingly complex environments for a wide range of applications. One such an application, that we will explore in this paper, is the use of autonomous robotic systems in socially challenging environments. In human-robot collaborative settings, robots require awareness of the current task as well as of the surrounding social environment. Consequently, they have to be able to make inferences in multiple levels of abstraction, to reason about and plan effectively, by combining task related actions and social interactions with humans \cite{7487760}.

In this paper we explore a social autonomous robotic system that performs a sorting game together with a small group of children, while it stays aware of the social-emotional states of the children in the same environment and keeps them emotionally engaged and positive. We aim for seamless social human-robot interaction. We consider existing theories to design socially appropriate robot strategies - high level categories for groups of behaviours - for a sustained child-robot interaction.

We use a planning-centric approach; prior to acting we use a planner to create a \emph{temporal plan} that achieves the goals of the robot (e.g. sort some toys according to predefined rules), while not breaking any of the constraints (e.g. leaving children in a prolonged negative emotional state). The benefit of using a planning-based approach is that by reasoning in advance, we can find good solutions that minimise time and pre-emptively interact with children to maintain their positive emotional states. The alternative is a reactive system that needs to stop fulfilling its task whenever a child is in a negative emotional state. Such a system has no guarantee when -- or indeed if -- it will finish its tasks. 

To create a robust \emph{plan}, we require a predictive model of the emotional states of the children. In particular how the emotions develop over time. Using this model, the planner can reason how the children's emotions are affected by the action (or inaction) of the robot. We present a predictive model based on the Pleasure-Arousal-Dominance~(PAD) emotional state model \cite{Mehrabian1996}, which was adapted to capture temporal features for the development of a dynamic interaction framework. 

In Section~\ref{sec:related work} we discuss the related work. In Section~\ref{sec:dynamic interaction framework} we present the predictive dynamic interaction framework, which is based on the PAD model. In Section~\ref{sec:planning system} we introduce the formalism of the planning system and some preliminary results. Then, we present initial steps towards the evaluation of our predictive model in Section~\ref{sec:evaluation}. We finish by presenting our conclusions and future directions in Section~\ref{sec:discussion and future work}.

%The developing child: What is special with still developing children?

%Goal: seamless social child-robot interaction in dynamic play environments: Considering child’s socio-%emotional state for robot’s decision making
 
\section{Related work}
\label{sec:related work}

\subsection{Planning-based social child-robot interaction}
\label{sec:Planning-based social child-robot interaction}

One of the focus points for the development of autonomous robots that interact with humans is the social intelligence of the system. Among the first attempts for the development of socially intelligent robots was that of Dauntenhahn's work \cite{dautenhahn1995getting}, who refers to robotic systems that collect mental and social experiences and based on this input mature over time. The robot considers user's behavioural, affective and mental states to provide appropriate responses for the evolution of the interaction loop. In this context, for the optimization of the interaction, robotic systems integrate planning and learning frameworks by taking into consideration human abilities and preferences \cite{7487760}, \cite{Kruse2013HumanawareRN}. More specifically, in the area of child-robot interaction, there is an increasing interest in the development of socially intelligent robots acting as learning companions for typically developing  children~ e.g. \cite{gordon2016affective} as well as therapeutic social agents for children on the autism spectrum e.g. \cite{esteban2017build}, \cite{Zheng2016}, and \cite{bernardini2013}. Despite the growing body of research on socially intelligent systems for child-robot interaction, the settings are usually well-defined and restricted, while the even more challenging area of child-robot interaction in dynamic play settings needs further investigation and development.

\subsection{Collaborative play and the importance of emotions}
\label{sec:Collaborative play and the importance of emotions}
This project uses a dynamic play setting for child-robot collaboration, in which the child and the robot share the same goal (e.g. sorting toys according to predefined rules) in the form of a guided activity \cite{Weisberg2015}. The importance of collaborative play for children's development has been previously highlighted in terms of children's developing cognitive and socio-emotional skills and the establishment of their intrinsic motivation for learning \cite{Lillard}. Based on the hypothesis that the development of these skills is more effective when the child is in an optimum affective state, previous research indicated a positive correlation of children's emotional competence with their concurrent and future social competences \cite{Denham2006}, as well as with the development of cognitive abilities \cite{Davis2013}. These findings indicate the importance of maintaining an optimum affective state for children during play. Towards this direction, this project incorporates approaches that focus on continuous input \cite{gunes2013categorical} and analysis of children's affective state.

\subsection{Models for affective states identification}
\label{sec:Models for affective states identification}

Emotional states in humans have been traditionally described by categorical or dimensional models. Categorical models such as the Differential Emotions Theory~(DET) \cite{Izard1992} and Ekman's theory of basic discrete emotional states \cite{Ekman1999} emphasize the existence of particular emotions that are assumed to have innate neural substrates, unique and universally recognized facial expressions and distinctive universals in antecedent events. On the other side, according to the dimensional approaches the emotion domain can be represented by a small number of continuous dimensions. Plutchik \cite{plutchik2001nature}, for example, suggested three dimensions: the emotional state, the intensity and the degree of similarity to other emotions. Recently, \cite{Barrett2017} suggested the theory of constructed  emotion, according to which an instance of emotion is constructed the same way that all other perceptions are constructed, using the same neuroanatomical principles for information flow within the brain, cancelling the distinct categorical nature of emotions. The dynamic nature of our setting requires a dimensional approach of emotions to depict changes of affective states over time.

\section{Towards a dynamic interaction framework}
\label{sec:dynamic interaction framework}

For the purpose of this project, we adopted the dimensional model of the Pleasure - Arousal - Dominance~(PAD) framework to detect, evaluate and predict users' emotional states. We used this model to develop a dynamic interaction framework that takes into consideration temporal features of affective development.

\subsection{The PAD model of emotions}
\label{sec:The PAD model of emotions}

Given the developmental nature of this project, which aims for long-term social human-robot interaction, we adopted the PAD model. The PAD model is a dimensional model that can be used to represent changes of emotional states over time. The PAD dimensional framework assumes that the dimensions of pleasure, arousal and dominance are necessary and sufficient to represent emotional states~\cite{Mehrabian1996}. They are described as follows:

\begin{enumerate}
 \item Pleasure-displeasure: Defined as positive versus negative affective states. Pleasure-displeasure corresponds to cognitive judgements of evaluation, with higher evaluations of stimuli being associated with greater pleasure induced by the stimuli;
 \item  Arousal-nonarousal: Defined in terms of level of mental alertness and physical activity;
 \item Dominance-submissiveness: Defined as a feeling of control and influence over one's surroundings and others versus feeling controlled or influenced by situations and others. 
\end{enumerate}

Adopting a model that considers the level of dominance, in addition to the traditionally used valence and arousal dimensions, represents children's emotional state more accurately. Especially given the collaborative nature of the settings in this project. For instance, both anger and anxiety arise from low-pleasure and high-arousal events. However, anger and anxiety are on opposite sides of the dominance dimension. The PAD framework has been previously used for the development of robotic systems in the context of social human-robot interaction \cite{6181762}. However, a recent systematic literature review \cite{charisi7745171} showed that there are a limited amount of studies that focus on developmental perspective for long-term sustained child-robot interaction in dynamic settings.

\subsection{Temporal considerations}
\label{sec:Temporal considerations}

Emotional processing is a dynamic phenomenon which is subject to stimuli such as external interventions from social agents. According to the generic timing hypothesis, an emotion is thought to come into being and develop through a recursive situation – attention – appraisal – response sequence \cite{doi:10.1177/1088868310395778}. The interventions distinguish between: antecedent-focused strategies that start operating early in a given iteration of the emotion-generative process, before response tendencies are fully activated; and response-focused strategies that start operating later on, after emotion response tendencies are more fully activated \cite{doi:10.1093/scan/nst116}. Based on this theory we hypothesize that temporal planning supports a balance of best performance in completing a task whilst maintaining appropriate emotions and engagement of the children.

In the context of this project, we focused on external interventions which are made by the robot. The robot starts with a perceived initial emotional state of the children. To maintain children's optimum emotional level, it applies an intervention / strategy to achieve user's reappraisal or attention deployment early in the emotion-generative trajectory, while monitoring the evolution of the user's emotional state.

\section{Planning system}
\label{sec:planning system}

We model our problem and domain files with PDDL~2.1~ \cite{fox2011}. This modelling language supports durative actions and temporal constraints. These features are necessary to capture the evolution of the emotional states over time. We will first define a temporal planning problem, followed by the model of the problem formulated in this paper.

\begin{definition}{Temporal Planning Problem Representation}
\label{def:classical planning problem representation}
We represent a temporal planning problem $C$ as $\mathit{P} = \langle F, I, A, G \rangle$ where $F$ is a set of atoms, $I$ is the set of clauses over $F$ representing the initial state, $G$ is a conjunction over $F$ that represents the goal that needs to be achieved, $A$ is a set of operators that affect the world. Every operator $a \in A$ has a precondition $\mathit{pre}(a)$ and a set of effects $\mathit{eff}(a)$. Each clause in the preconditions and effects are annotated with a \textit{temporal constraint}; A precondition clause must either hold: at the beginning of the action, at the end of the action, or during the entire duration. Effects are applied either at the beginning or end of an action.
\end{definition}

\begin{figure}[H]
\scriptsize
\begin{Verbatim}[tabsize=1]
(define (problem squirrel_emotion_problem)
(:domain squirrel_emotion)
(:objects
	toy1 toy2 toy3 - object
	box1 - box
	kenny - robot
	kenny_wp toy1_wp toy2_wp toy3_wp box1_wp - waypoint
)
(:init
	(not_busy)
	(robot_at kenny kenny_wp)
	(box_at box1 box1_wp)
	(object_at toy1 toy1_wp)
	(object_at toy2 toy2_wp)
	(object_at toy3 toy3_wp)
	(gripper_empty kenny)

	(= (pleasure c1) 0.4)
	(= (arousal c1) 0.4)
	(= (dominance c1) 0.45)

	(= (pleasure c2) 1.0)
	(= (arousal c2) 1.0)
	(= (dominance c2) 1.0)

	(= (pleasure c3) 0.83)
	(= (arousal c3) 0.98)
	(= (dominance c3) 0.6)
)
(:goal (and
	(in_box box1 toy1)
	(in_box box1 toy2)
	(in_box box1 toy3)
)))
\end{Verbatim}
\caption{The problem.}
\label{fig:pddl_problem}
\end{figure}
%\vspace{-4mm}

\begin{figure}[ht!]
\scriptsize
\begin{Verbatim}[tabsize=1]
(define (domain squirrel_emotion)
(:requirements ...)
(:types robot child	waypoint box object)
(:functions
	(pleasure ?c - child)
	(arousal ?c - child)
	(dominance ?c - child)
)

(:constants c1 c2 c3 - child)

(:predicates
	(robot_at ?v - robot ?wp - waypoint)
	(object_at ?o - object ?wp - waypoint)
	(box_at ?b - box ?wp - waypoint)
	(classified ?o - object)
	(in_box ?b - box ?o - object)
	(holding ?v - robot ?o - object)
	(gripper_empty ?v - robot)
	(not_busy)
)

(:durative-action accommodate-distress
	:parameters (?c - child)
	:duration (<= ?duration 30)
	:condition (and
		(over all (< (pleasure ?c) 1))
		(over all (< (arousal ?c) 0))
		(at start (< (pleasure ?c) 0.5))
		(at start (> (arousal ?c) 0.5))
		(at start (> (dominance ?c) 0.5))
		(at start (not_busy))
	)
	:effect (and
		(at start (not (not_busy)))
		(at end (not_busy))
		(at end (increase (pleasure ?c) 
		           (* ?duration 0.01)))
		(at end (decrease (arousal ?c) 
		          (* ?duration 0.02)))
	)
)

(:durative-action improve-distress ...)
(:durative-action accommodate-sadness ...)
(:durative-action improve-sadness ...)
(:durative-action improve-boredom ...)
(:durative-action maintain-happiness ...)
(:durative-action improve-introvert ...)

(:durative-action kid_give
	:parameters (...)
	:duration (= ?duration 60)
	:condition (and
		(over all (robot_at ?v ?robot_wp))
        (at start (gripper_empty ?v))
		(at start (object_at ?o ?object_wp))
		(at start (not_busy))
		(over all (<= (pleasure ?c) 1))
		(over all (<= (arousal ?c) 1))
		(over all (<= (dominance ?c) 1))
	)
	:effect (and
		(at start (not (not_busy)))
		(at end (not_busy))
		(at start (not (gripper_empty ?v)))
		(at end (holding ?v ?o))
		(at start (not (object_at ?o ?object_wp)))
		(at end (increase (pleasure ?c) (* ?duration 0.005)))
		(at end (increase (arousal ?c) (* ?duration 0.005)))
		(at end (increase (dominance ?c) (* ?duration 0.005)))
	)
)

(:durative-action move ...)
(:durative-action classify ...)
(:durative-action pickup ...)
(:durative-action tidy ...)
)
\end{Verbatim}
\caption{Fragment of the PDDL domain.}
\label{fig:pddl_domain}
\end{figure}

\subsection{Modelling the planning problem}

Using the PAD model described in Section~\ref{sec:dynamic interaction framework} we created a planning model that encapsulates the children's emotional state and its evolution of time. In this paper we use a case study from the EU project SQUIRREL~\footnote{http://www.squirrel-project.eu/}; The robot is tasked with sorting a set of toys while at the same time collaborating with three children that are active in the same area. We assume we know the initial emotional state of the children and we have an array of sensors to monitor the children's emotional state during execution as described in section ~\ref{sec:evaluation}.

The PDDL Domain is listed in Figure~\ref{fig:pddl_domain}, most actions have been abbreviated due to space constraints. Children's emotional states are encoded using a triplet of functions that correspond to the three domains of the PAD-model: \textit{pleasure}, \textit{arousal}, \textit{dominance}. All actions in the domain affect the emotional state of each child. It is assumed that robot's task-related actions usually have less effect on the emotional state of the children and generally tend to lower \textit{pleasure} which will eventually lead to boredom. 

Social-emotional actions, like \textit{accommodate-distress}, have more effect on the children's emotional states as they tend to interact with children directly and not contribute to the overall task. We present three strategies that can be used by the robot to alter the children's emotions. There are:

\begin{enumerate}
 \item Accommodate: The robot gives the time for the child to familiarize him/herself to the new situation.
 \item Maintain: The robot has an interactive role to maintain the positive state of the child.
 \item Improve: The robot initiates and actively applies strategies to trigger a change from a neutral to a positive state.
\end{enumerate}

Each strategy contains a set of various behaviours for execution.One action does both at the same time; the action \textit{kid-give} has the robot ask a child to give it an item. This action contributes to the task and improves the emotional state of the child that helps the robot.
The effect of these strategies depend on the current emotional state of the child. For instance, if a child is in a very positive emotional state (e.g. Pleasure, Arousal, and Dominance are both high) then applying the Improve strategy will not affect the emotional state much. However, if the child is sad (e.g. Pleasure and Arousal are low, but Dominance is high) then applying the improve strategy will have a noticeable, positive, effect on the child's emotion.

We have modelled four separate emotions we can detect in children, \textit{Distress}, \textit{Sadness}, \textit{Boredom}, and \textit{Happiness}. The relevant relations between these emotions and the PAD levels are depicted in Table~\ref{table:pad to emotions}.

\begin{table}[ht]
\centering
\begin{tabular}{|l|l|l|l|}
\hline
           & \multicolumn{3}{c|}{PAD values}  \\ \hline
Emotions   & Pleasure & Arousal   & Dominance \\ \hline
Distress   & Low      & High      & High      \\ \hline
Sadness    & Low      & Low       & High      \\ \hline
Boredom    & Low      & Low       & Low       \\ \hline
Happiness  & High     & High/Low  & High      \\ \hline
\end{tabular}
\caption{Relations between emotions and the PAD values.}
\label{table:pad to emotions}
\end{table}

The effects of the actions on the emotional states are listed in Table~\ref{table:actions effects on emotions}. In our domain we model the three domains of the PAD model using numerical values. We limit the range of these values between -1 and 1, where 1 is the highest value and corresponds to the \textit{high} value. 0 is considered to be \textit{low}. We do not want any of these domains to become \textit{low} during planning execution, so the robot is unable to execute any task-related action until the emotional states of all children are not negative.

\begin{table}[ht]
\centering
\begin{tabular}{|l||c|c|c||c|c|c||c|c|c|}
\hline
 & \multicolumn{3}{c||}{\textbf{Accommodate}} & \multicolumn{3}{c||}{\textbf{Maintain}} & \multicolumn{3}{c|}{\textbf{Improve}}  \\ \hline
\textbf{Emotions}   & \textit{P} & \textit{A} & \textit{D} & \textit{P} & \textit{A} & \textit{D} & \textit{P} & \textit{A} & \textit{D} \\ \hline\hline
Distress   & + &-- & 0 & 0 & 0 & 0 & ++& --& 0 \\ \hline
Sadness    & + & 0 & 0 & 0 & 0 & 0 & ++& 0 & 0 \\ \hline
Boredom    & 0 & 0 & 0 & 0 & 0 & 0 & + & + & + \\ \hline
Happiness  & 0 & 0 & 0 & 0 & - & 0 & 0 & 0 & 0 \\ \hline
\end{tabular}
\caption{Effect of the possible actions on the state of the children.}
\label{table:actions effects on emotions}
%\vspace{-10mm}
\end{table}

An example planning problem is listed in Figure~\ref{fig:pddl_problem}. We define the emotional state of each of the three children (c1, c2, and c3). The child c1 is bordering boredom, c2 is very happy, and c3 is satisfied but not very active. The goal is to store away the three toys (toy1, toy2, and toy3) in the provided box.

We use the temporal planner POPF \cite{coles2010} to solve this planning problem. A solution is depicted in Figure~\ref{fig:pddl_plan}.

\begin{figure*}[!tb]
\footnotesize{
\begin{minipage}[b]{0.32\textwidth}
  \centerline{\includegraphics[width=\linewidth]{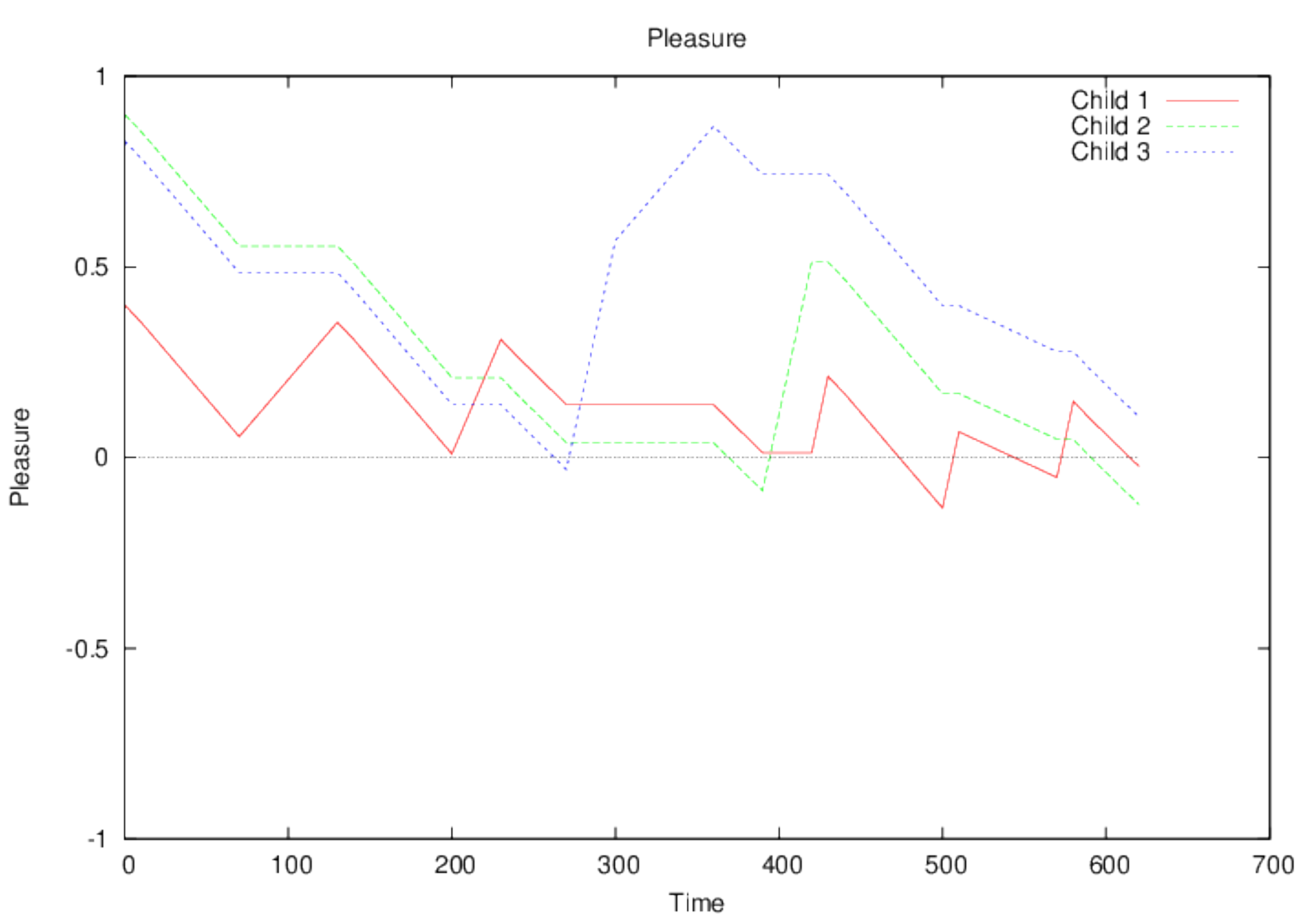}}
  \centering{(a) Pleasure}
\end{minipage}
\begin{minipage}[b]{0.32\textwidth}
  \centerline{\includegraphics[width=\linewidth]{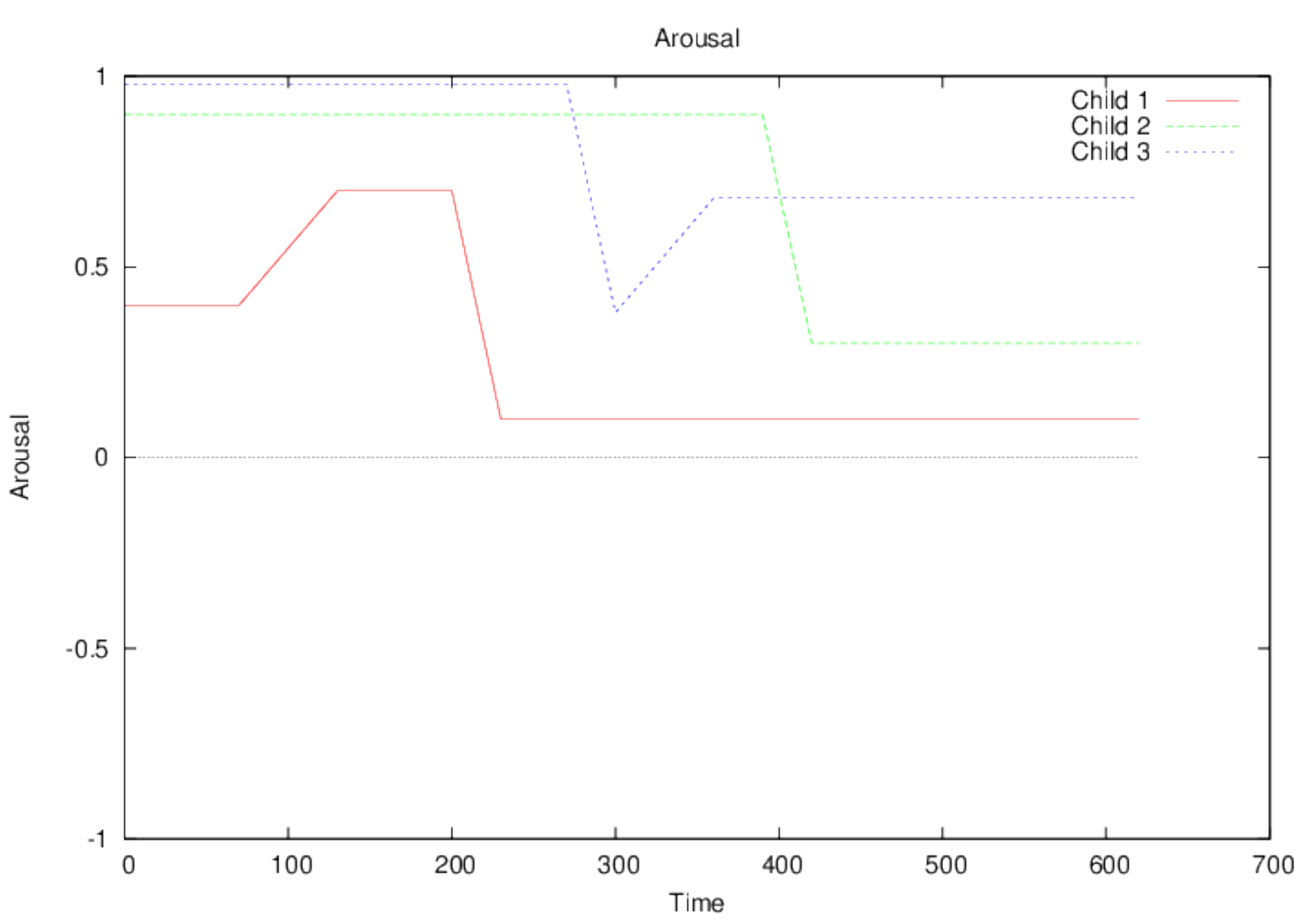}}
  \centering{(b) Arousal}
\end{minipage}
\begin{minipage}[b]{0.32\textwidth}
  \centerline{\includegraphics[width=\linewidth]{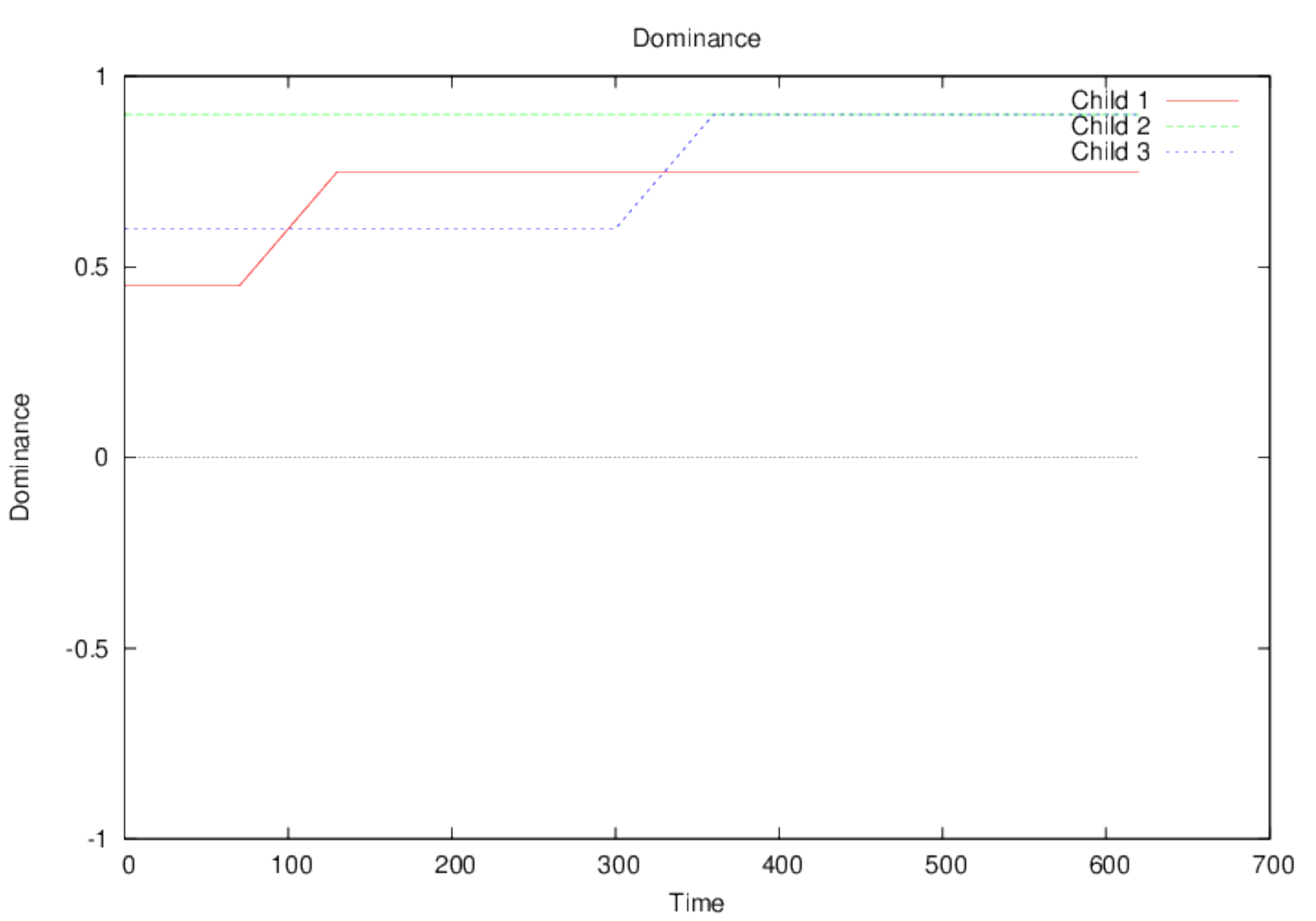}}
  \centering{(c) Dominance}
\end{minipage}
}
%\vspace{-0.1cm}
\caption{The evolution of the emotions of the children over time.}
\label{fig:PAD evolution}
\end{figure*}

The evolution of the Pleasure, Arousal, and Dominance are depicted in Figure~\ref{fig:PAD evolution}. The planner aims to minimise the time it takes to complete the task; It allows the emotional states of the children to border the acceptable and keep it there.

\begin{figure}[h!]
\scriptsize
\begin{Verbatim}[tabsize=1]
0.000: (move kenny kenny_wp toy1_wp)  [10.000]
10.001: (classify kenny toy1 toy1_wp)  [60.000]
70.002: (kid_give c1 kenny toy1 toy1_wp toy1_wp)  [60.000]
130.003: (move kenny toy1_wp toy3_wp)  [10.000]
140.004: (classify kenny toy3 toy3_wp)  [60.000]
200.005: (accomodate-distress c1)  [30.000]
230.006: (move kenny toy3_wp box1_wp)  [10.000]
240.007: (tidy kenny toy1 box1 box1_wp)  [30.000]
270.008: (improve-distress c3)  [30.000]
300.009: (kid_give c3 kenny toy3 box1_wp toy3_wp)  [60.000]
360.010: (tidy kenny toy3 box1 box1_wp)  [30.000]
390.011: (improve-distress c2)  [30.000]
420.012: (improve-sadness c1)  [10.000]
430.013: (move kenny box1_wp toy2_wp)  [10.000]
440.014: (classify kenny toy2 toy2_wp)  [60.000]
500.015: (improve-sadness c1)  [10.000]
510.016: (pickup kenny toy2 toy2_wp)  [60.000]
570.017: (improve-sadness c1)  [10.000]
580.018: (move kenny toy2_wp box1_wp)  [10.000]
590.019: (tidy kenny toy2 box1 box1_wp)  [30.000]
\end{Verbatim}
\caption{A possible plan.}
\label{fig:pddl_plan}
\vspace{-5mm}
\end{figure}

\section{Preliminary Evaluation}
\label{sec:evaluation}

We present an empirical pilot study for the evaluation of the proposed Dynamic Interaction Framework; in this study we detect and interpret children's arousal level as an indicator of their task engagement. We define a threshold of arousal level. The robot performs a strategy to improve children's arousal level by executing an unexpected behaviour.

\subsection{Setting}

\begin{figure*}[!tb]
\footnotesize{
\begin{minipage}[b]{0.32\textwidth}
  \centerline{\includegraphics[width=\linewidth]{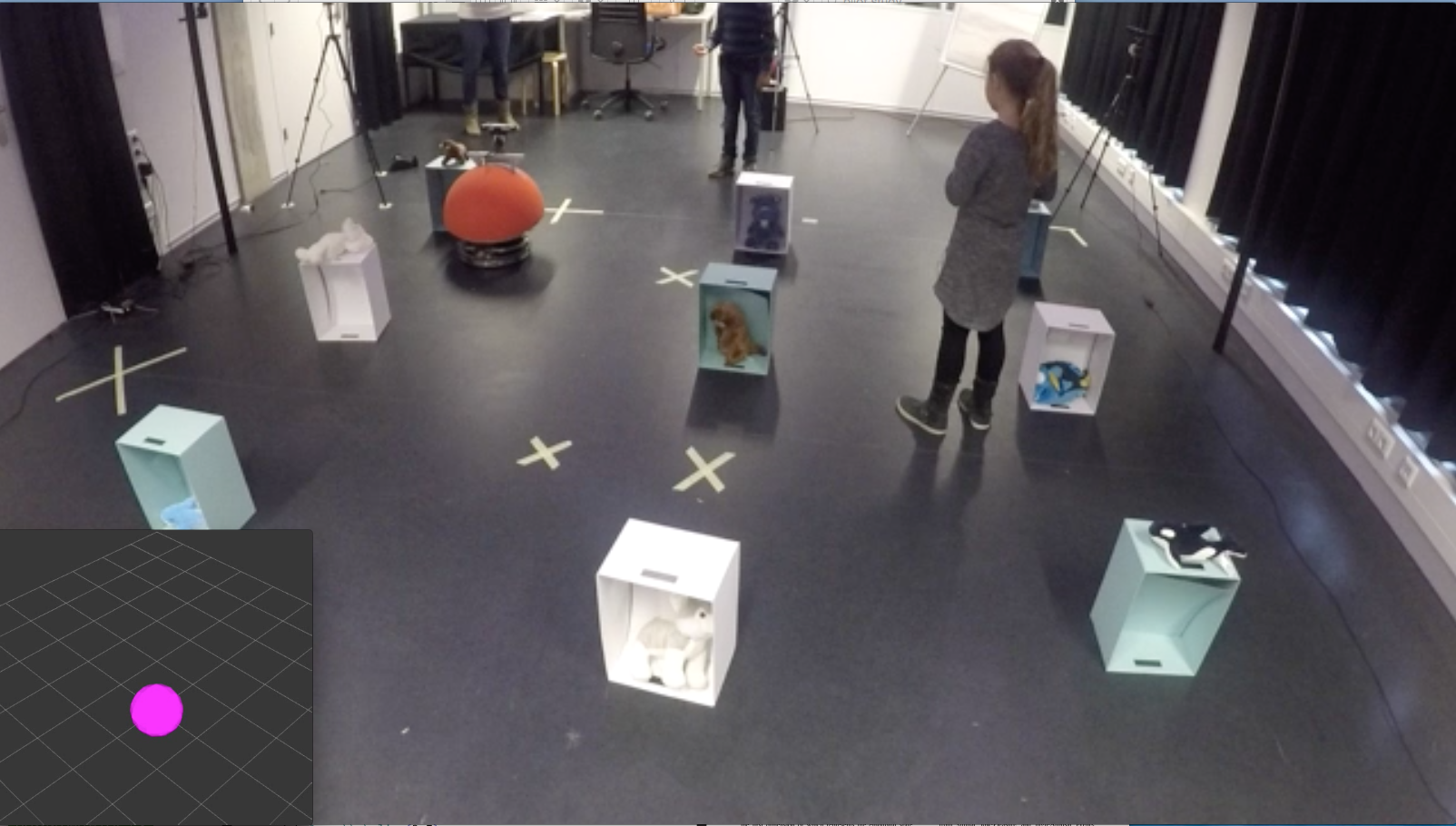}}
  \centering{(a) }
\end{minipage}
\begin{minipage}[b]{0.32\textwidth}
  \centerline{\includegraphics[width=\linewidth]{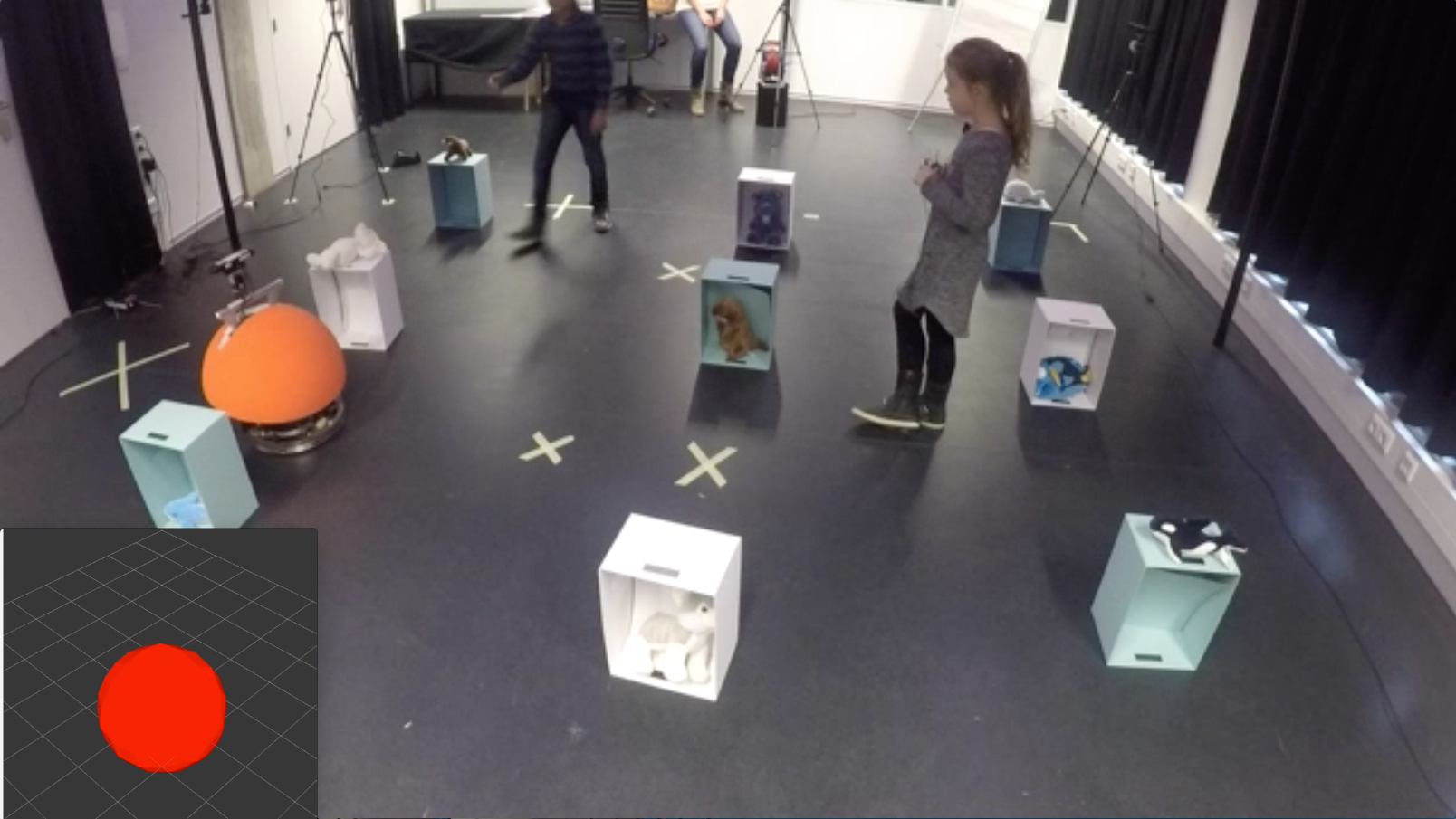}}
  \centering{(b) }
\end{minipage}
\begin{minipage}[b]{0.32\textwidth}
  \centerline{\includegraphics[width=\linewidth]{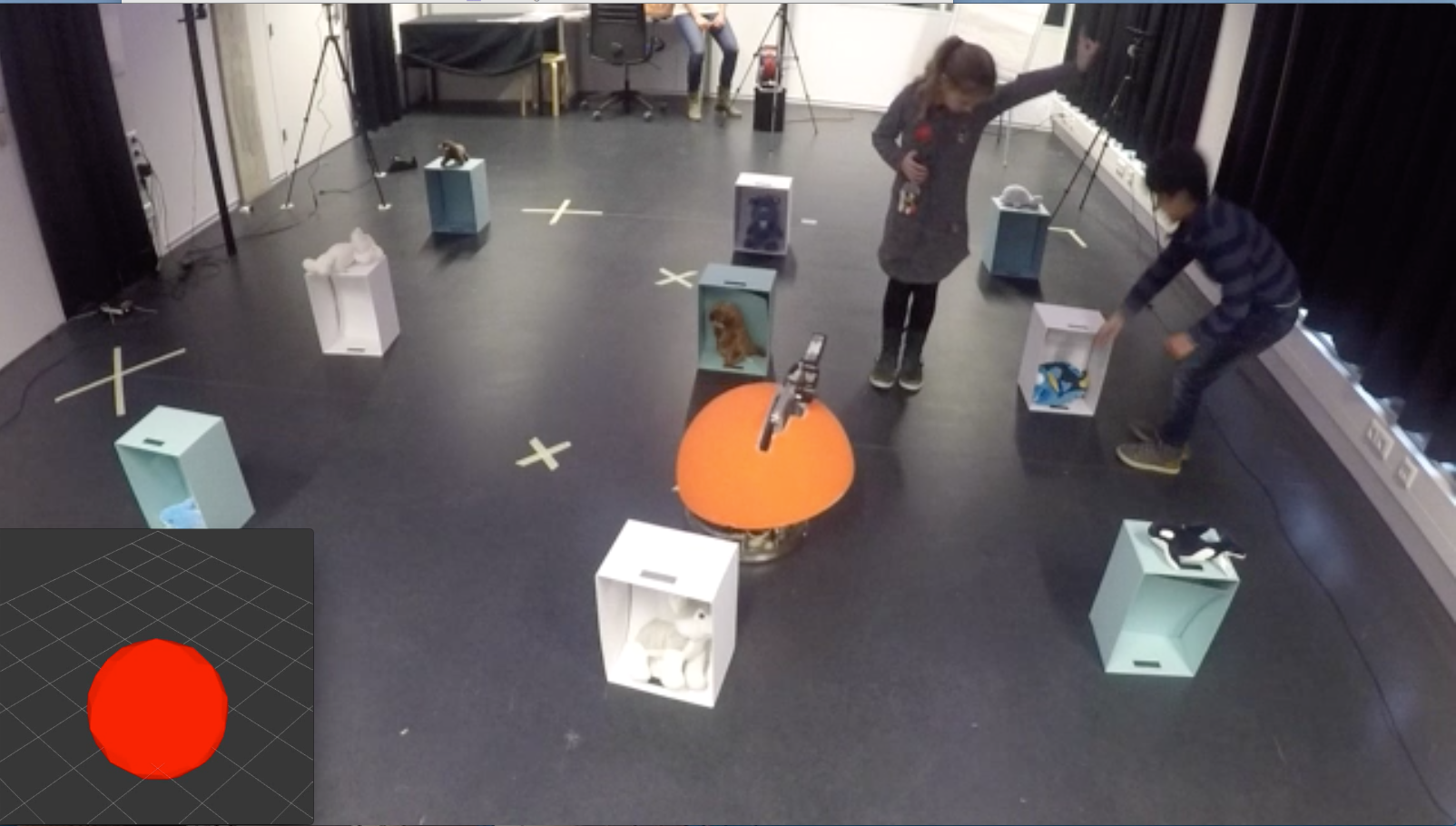}}
  \centering{(c) }
\end{minipage}
}
%\vspace{-0.1cm}
\caption{Observations on children's behaviours and visualised arousal levels, at the left corner, a small circle indicates low-level of arousal, a large circle indicates high-level of arousal.}
\label{fig:observations}
\end{figure*}

We conducted two sessions. In each session two children aged 8-9 years played together with a non-humanoid robot with the aim to sort a set of toys according to predefined rules. In order to record individual behaviours (including speech, pose, and gestures) and avoid occlusions, we used individual Lapel microphones and 4 Kinects. We did not employ a speaker identification system in this study; We relied on individual recordings that are not easily applicable to children in the wild. The speaker identification issue is considered in future work.

\subsection{Speech Emotion Detection System}
In a dynamic play environment, we assume that children frequently show occlusions between their motions and it would be challenging to track their faces constantly. Hence, we used a speech emotion recognition system to monitor their affective states. In this pilot study, we focused only on measurement of children's arousal level. To this end, we have developed a deep multi-task learning based speech emotion recognition model using aggregated corpora that provides better generalisation. The model has two layers of Long-Short-Term-Memory~(LSTM) with 128 cells. Details of the method and used corpora can be found in \cite{kim2017interspeech}. The unweighed accuracy on the arousal dimension (low, high levels) was $82$\%.

The system consists of three modules: voice activity detector, feature extractor, and classifier. For robustness in a noisy environment, we adopt Gaussian Mixture Models classifying frames with a length of 20ms into speech and non-speech frames. Then, consecutive speech frames bridged by a short silence (shorter than a half sec.) but segmented by a long silence (longer than a half sec.) forms an utterance to classify. We only classify sufficiently long utterances (longer than 1 sec.). Next, manually-engineered feature vectors are extracted from an utterance (See the details in 
\cite{kim2017interspeech}. Lastly, the trained LSTM network estimates the probabilistic distribution of the two classes.

\subsection{Results}
We classified utterances extracted from each recording of a child to analyse their arousal level. Table~\ref{tab:emotionresult} summarises the classified states. As shown, we found more utterances with the high level of arousal in session 1 than those in session 2, which is aligned with our behavioural observations. In addition, we observed how the robot's strategy of unexpected behaviour (i.e. the robot did not follow the verbal command of the child) affected arousal states of children. Figure \ref{fig:observations} presents examples. We detected arousal states regardless of who speaks. First, (a) shows there was no high-level of arousal when the robot behaved as expected by the children. However, (b) and (c) showed high-level of arousal when the robot demonstrated unexpected behaviours, which indicate the efficiency of the strategy in the specific context.

\begin{table}[!t]
\centering
%\vspace{10px}
\begin{tabular}{lllllll}
\hline
Session   & \multicolumn{2}{c}{Child A} & \multicolumn{2}{c}{Child B} & \multicolumn{2}{c}{Average}\\
\hline
			& High 	& Low 			& High 	& Low 		 & High	& Low\\ \cline{2-7}
1			& $72.4$	& $27.6$	& $70.2$	& $29.8$ & $71.2$ 	& $28.8$\\
2			& $46.8$	& $53.2$	& $34.6$	& $65.4$ & $44.4$	& $55.6$\\	
\hline
\end{tabular}
\caption{Distribution (\%) of emotional states: high and low arousal}\label{tab:emotionresult}
\vspace{-10mm}
\end{table}

%Any qualitative analysis based on your observations? For example, we observed the first two children showed higher arousal compared to the last two children.
\section{Discussion and Future Work}
\label{sec:discussion and future work}

This paper describes the initial steps towards the design of a planning based robotic system for social child-robot interaction in a play environment. We have proposed a Dynamic Interaction Framework based on the existing PAD model of emotions for social HRI. The robot uses a planning to create plans that complete tasks while being socially aware and executing specific strategies to keep the interacting children positive and engaged. 

The temporal model we created for this scenario includes task-related actions and social-emotional actions that have the robot interact with the children directly to improve their emotional state. By creating a plan in advance we can pre-emptively improve children's emotional states and finish the task in a good time.

Finally, we have presented a pilot study; we evaluated part of the proposed Dynamic Interaction Framework as well as the strategy of robot's unexpected behaviour. We demonstrated that the framework is applicable in real settings and the strategy has a positive impact on children's arousal level.

The proposed Dynamic Interaction Framework aims to support child-robot interaction in dynamic play settings but it has some limitations. One of the major challenges relates to temporal considerations. While, the framework takes into account timing aspects for the execution of a specific strategy, due to the complexity of the dynamic setting this is challenging to be accurate enough during the execution.

In future work, we intend to empirically investigate temporal aspects of robot's behaviour in play environments and their effectiveness. In addition, we aim to integrate further modalities for the identification of children's emotional states and engagement level. By further developing the Dynamic Interaction Framework for planning based robotic systems, we aim to improve its transferability in socially complex settings such as children's play environments. 

%\section*{Acknowledgments}

%\newpage

\bibliographystyle{plainnat}
\bibliography{references-nourl}

\begin{thebibliography}{24}
\providecommand{\natexlab}[1]{#1}
\providecommand{\url}[1]{\texttt{#1}}
\expandafter\ifx\csname urlstyle\endcsname\relax
  \providecommand{\doi}[1]{doi: #1}\else
  \providecommand{\doi}{doi: \begingroup \urlstyle{rm}\Url}\fi

\bibitem[Barrett(2017)]{Barrett2017}
L.~F. Barrett.
\newblock The theory of constructed emotion: an active inference account of
  interoception and categorization.
\newblock \emph{Social cognitive and affective neuroscience}, 12\penalty0
  (1):\penalty0 1--23, 2017.

\bibitem[Bernardini and Porayska-Pomsta(2013)]{bernardini2013}
S.~Bernardini and K.~Porayska-Pomsta.
\newblock \emph{Planning-Based Social Partners for Children with Autism},
  volume N/A, pages 362--370.
\newblock AAAI Press, n/a edition, 2013.

\bibitem[Charisi et~al.(2016)Charisi, Davison, Reidsma, and
  Evers]{charisi7745171}
V.~Charisi, D.~Davison, D.~Reidsma, and V.~Evers.
\newblock Evaluation methods for user-centered child-robot interaction.
\newblock In \emph{2016 25th IEEE International Symposium on Robot and Human
  Interactive Communication (RO-MAN)}, pages 545--550, Aug 2016.
\newblock \doi{10.1109/ROMAN.2016.7745171}.

\bibitem[Coles et~al.(2010)Coles, Coles, Fox, and Long]{coles2010}
A.~J. Coles, A.~I. Coles, M.~Fox, and D.~Long.
\newblock Forward-chaining partial-order planning.
\newblock In \emph{Proceedings of the Twentieth International Conference on
  Automated Planning and Scheduling (ICAPS-10)}, May 2010.

\bibitem[Dautenhahn(1995)]{dautenhahn1995getting}
K.~Dautenhahn.
\newblock Getting to know each other—artificial social intelligence for
  autonomous robots.
\newblock \emph{Robotics and autonomous systems}, 16\penalty0 (2-4):\penalty0
  333--356, 1995.

\bibitem[Davis and Levine(2013)]{Davis2013}
E.~L. Davis and L.~J. Levine.
\newblock Emotion regulation strategies that promote learning: Reappraisal
  enhances children’s memory for educational information.
\newblock \emph{Child Development}, 84\penalty0 (1):\penalty0 361--374, 2013.

\bibitem[Denham(2006)]{Denham2006}
S.~A. Denham.
\newblock Social-emotional competence as support for school readiness: What is
  it and how do we assess it?
\newblock \emph{Early education and development}, 17\penalty0 (1):\penalty0
  57--89, 2006.

\bibitem[Ekman(1999)]{Ekman1999}
P.~Ekman.
\newblock Facial expressions.
\newblock \emph{Handbook of cognition and emotion}, 16:\penalty0 301--320,
  1999.

\bibitem[Esteban et~al.(2017)Esteban, Baxter, Belpaeme,
  et~al.]{esteban2017build}
P.~G. Esteban, P.~Baxter, T.~Belpaeme, et~al.
\newblock How to build a supervised autonomous system for robot-enhanced
  therapy for children with autism spectrum disorder.
\newblock \emph{Paladyn, Journal of Behavioral Robotics}, 8\penalty0 (1), 2017.

\bibitem[Fox and Long(2011)]{fox2011}
M.~Fox and D.~Long.
\newblock {PDDL2.1:} an extension to {PDDL} for expressing temporal planning
  domains.
\newblock \emph{CoRR}, abs/1106.4561, 2011.

\bibitem[Gordon et~al.(2016)Gordon, Spaulding, Westlund,
  et~al.]{gordon2016affective}
G.~Gordon, S.~Spaulding, J.~Kory Westlund, et~al.
\newblock Affective personalization of a social robot tutor for children’s
  second language skills.
\newblock In \emph{Thirtieth AAAI Conference on Artificial Intelligence}, 2016.

\bibitem[Gunes and Schuller()]{gunes2013categorical}
H.~Gunes and B.~Schuller.
\newblock Categorical and dimensional affect analysis in continuous input:
  Current trends and future directions.

\bibitem[Hayes and Scassellati(2016)]{7487760}
B.~Hayes and B.~Scassellati.
\newblock Autonomously constructing hierarchical task networks for planning and
  human-robot collaboration.
\newblock In \emph{2016 IEEE International Conference on Robotics and
  Automation (ICRA)}, pages 5469--5476, May 2016.
\newblock \doi{10.1109/ICRA.2016.7487760}.

\bibitem[Izard(1992)]{Izard1992}
C.~E. Izard.
\newblock Basic emotions, relations among emotions, and emotion-cognition
  relations.
\newblock 1992.

\bibitem[Kim et~al.(2017)Kim, Englebienne, Truong, and
  Evers]{kim2017interspeech}
J.~Kim, G.~Englebienne, K.~P. Truong, and V.~Evers.
\newblock Towards speech emotion recognition ``in the wild'' using aggregated
  corpora and deep multi-task learning.
\newblock In \emph{Proceedings of INTERSPEECH}, page To be appeared, 2017.

\bibitem[Kruse et~al.(2013)Kruse, Pandey, Alami, and
  Kirsch]{Kruse2013HumanawareRN}
T.~Kruse, A.~K. Pandey, R.~Alami, and A.~Kirsch.
\newblock Human-aware robot navigation: A survey.
\newblock \emph{Robotics and Autonomous Systems}, 61:\penalty0 1726--1743,
  2013.

\bibitem[Lillard(2015)]{Lillard}
A.~S. Lillard.
\newblock \emph{The Development of Play}.
\newblock John Wiley, Inc., 2015.
\newblock ISBN 9781118963418.
\newblock \doi{10.1002/9781118963418.childpsy211}.

\bibitem[Mehrabian(1996)]{Mehrabian1996}
A.~Mehrabian.
\newblock Pleasure-arousal-dominance: A general framework for describing and
  measuring individual differences in temperament.
\newblock \emph{Current Psychology}, 14\penalty0 (4):\penalty0 261--292, 1996.
\newblock ISSN 1936-4733.
\newblock \doi{10.1007/BF02686918}.

\bibitem[Park et~al.(2011)Park, Kim, et~al.]{6181762}
J.~W. Park, W.~H. Kim, et~al.
\newblock How to completely use the pad space for socially interactive robots.
\newblock In \emph{2011 IEEE International Conference on Robotics and
  Biomimetics}, pages 3005--3010, Dec 2011.

\bibitem[Plutchik(2001)]{plutchik2001nature}
R.~Plutchik.
\newblock The nature of emotions human emotions have deep evolutionary roots, a
  fact that may explain their complexity and provide tools for clinical
  practice.
\newblock \emph{American scientist}, 89\penalty0 (4):\penalty0 344--350, 2001.

\bibitem[Schönfelder et~al.(2014)Schönfelder, Kanske, Heissler, and
  Wessa]{doi:10.1093/scan/nst116}
S.~Schönfelder, P.~Kanske, J.~Heissler, and M.~Wessa.
\newblock Time course of emotion-related responding during distraction and
  reappraisal.
\newblock \emph{Social Cognitive and Affective Neuroscience}, 9\penalty0
  (9):\penalty0 1310, 2014.
\newblock \doi{10.1093/scan/nst116}.

\bibitem[Sheppes and Gross(2011)]{doi:10.1177/1088868310395778}
G.~Sheppes and J.~J. Gross.
\newblock Is timing everything? temporal considerations in emotion regulation.
\newblock \emph{Personality and Social Psychology Review}, 15\penalty0
  (4):\penalty0 319--331, 2011.

\bibitem[Weisberg et~al.(2015)Weisberg, Kittredge, Hirsh-Pasek,
  et~al.]{Weisberg2015}
D.~S. Weisberg, A.~K. Kittredge, K.~Hirsh-Pasek, et~al.
\newblock Making play work for education.
\newblock \emph{Phi Delta Kappan}, 96\penalty0 (8):\penalty0 8--13, 2015.
\newblock \doi{10.1177/0031721715583955}.

\bibitem[Zheng et~al.(2016)Zheng, Warren, Weitlauf, et~al.]{Zheng2016}
Z.~Zheng, Z.~Warren, A.~Weitlauf, et~al.
\newblock Brief report: Evaluation of an intelligent learning environment for
  young children with autism spectrum disorder.
\newblock \emph{Journal of Autism and Developmental Disorders}, 46\penalty0
  (11):\penalty0 3615--3621, 2016.
\newblock \doi{10.1007/s10803-016-2896-0}.

\end{thebibliography}

\end{document}